\documentclass{article}

\usepackage[preprint]{neurips_2019}

\usepackage[T1]{fontenc}    
\usepackage{hyperref}       
\usepackage{url}            
\usepackage{booktabs}       
\usepackage{amsfonts}       
\usepackage{nicefrac}       
\usepackage{microtype}      
\usepackage{float}
\usepackage{array}
\usepackage[utf8]{inputenc}
\usepackage{graphicx}

\title{"Hinglish" Language - Modeling a Messy Code-Mixed Language}

\author{
  Vivek Kumar Gupta\\
  \texttt{vkgupta@stanford.edu} \\
}

\begin{document}
\maketitle
\begin{abstract}
  With a sharp rise in fluency and users of "Hinglish" in linguistically diverse country, India, it has increasingly become important to analyze social content written in this language in platforms such as Twitter, Reddit, Facebook. This project focuses on using deep learning techniques to tackle a classification problem in categorizing social content written in Hindi-English into Abusive, Hate-Inducing and Not offensive categories. We utilize bi-directional sequence models with easy text augmentation techniques such as synonym replacement, random insertion, random swap, and random deletion to produce a state of the art classifier that outperforms the previous work done on analyzing this dataset.   
\end{abstract}

\section{Introduction}

Hinglish is a linguistic blend of Hindi (very widely spoken language in India) and English (an associate language of urban areas) and is spoken by upwards of 350 million people in India. While the name is based on the Hindi language, it does not refer exclusively to Hindi, but is used in India, with English words blending with Punjabi, Gujarati, Marathi and Hindi. Sometimes, though rarely, Hinglish is used to refer to Hindi written in English script and mixing with English words or phrases. This makes analyzing the language very interesting. Its rampant usage in social media like Twitter, Facebook, Online blogs and reviews has also led to its usage in delivering hate and abuses in similar platforms. We aim to find such content in the social media focusing on the tweets. Hypothetically, if we can classify such tweets, we might be able to detect them and isolate them for further analysis before it reaches public. This will a great application of AI to the social cause and thus is motivating. 
An example of a simple, non offensive message written in Hinglish could be:

\begin{center}
  \texttt{"Why do you waste your time with \textit{<redacted content>}. Aapna ghar sambhalta nahi(\textit{<redacted content>}). Chale dusro ko basane..!!"}
\end{center}

The second part of the above sentence is written in Hindi while the first part is in English. Second part calls for an action to a person to bring order to his/her home before trying to settle others.

\subsection{Modeling challenges}

From the modeling perspective there are couple of challenges introduced by the language and the labelled dataset. Generally, Hinglish follows largely fuzzy set of rules which evolves and is dependent upon the users preference. It doesn't have any formal definitions and thus the rules of usage are ambiguous. Thus, when used by different users the text produced may differ. Overall the challenges posed by this problem are:
\begin{itemize}
    \item \textbf{Geographical variation}: Depending upon the geography of origination, the content may be be highly influenced by the underlying region. 
    \item  \textbf{Language and phonetics variation}: Based on a  census in 2001, India has 122 major languages and 1599 other languages. The use of Hindi and English in a code switched setting is highly influenced by these language.
    \item  \textbf{No grammar rules}: Hinglish has no fixed set of grammar rules. The rules are inspired from both Hindi and English and when mixed with slur and slang  produce large variation.
    \item  \textbf{Spelling variation}: There is no agreement on the spellings of the words which are mixed with English. For example to express \texttt{love}, a code mixed spelling, specially when used social platforms might be \texttt{pyaar}, \texttt{pyar} or \texttt{pyr}.
    \item  \textbf{Dataset}: Based on some earlier work, only available labelled dataset had 3189 rows of text messages of average length of 116 words and with a range of {1, 1295}. Prior work addresses this concern by using Transfer Learning on an architecture learnt on  about 14,500 messages with an accuracy of 83.90. We addressed this concern using data augmentation techniques applied on text data. 
\end{itemize}

\section{Related Work}

\subsection{Transfer learning based approaches}
Mathur et al. in their paper for detecting offensive tweets proposed a Ternary Trans-CNN model where they train a model architecture comprising of 3 layers of Convolution 1D having filter sizes of 15, 12 and 10 and kernel size of 3 followed by 2 dense fully connected layer of size 64 and 3. The first dense FC layer has  ReLU activation while the last Dense layer had Softmax activation. They were able to train this network on a parallel English dataset provided by Davidson et al. The authors were able to achieve Accuracy of 83.9\%, Precision of 80.2\%, Recall of 69.8\%.

The approach looked promising given that the dataset was merely 3189 sentences divided into three categories and thus we replicated the experiment but failed to replicate the results. The results were poor than what the original authors achieved. But, most of the model hyper-parameter choices where inspired from this work. 

\subsection{Hybrid models}
In another localized setting of Vietnamese language, Nguyen et al. in 2017 proposed a Hybrid multi-channel CNN and LSTM model where they build feature maps for Vietnamese language using CNN to capture shorterm dependencies and LSTM to capture long term dependencies and concatenate both these feature sets to learn a unified set of features on the messages. These concatenated feature vectors are then sent to a few fully connected layers. They achieved an accuracy rate of 87.3\% with this architecture. 

\section{Dataset and Features}

We used dataset, HEOT obtained from one of the past studies done by Mathur et al. where they annotated a set of cleaned tweets obtained from twitter for the conversations happening in Indian subcontinent. A labelled dataset for a corresponding english tweets were also obtained from a study conducted by Davidson et al. This dataset was important to employ Transfer Learning to our task since the number of labeled dataset was very small. 
Basic summary and examples of the data from the dataset are below:

\begin{table}[H]
  \caption{Annotated Data set}
  \label{labelleddataset}
  \centering
  \begin{tabular}{lll}
    \toprule
    \multicolumn{2}{c}{Hinglish and English Data}                   \\
    \cmidrule(r){1-3}
    Label     & HOT     & English \\
    \midrule
    Non-Offensive & 1121  & 7274     \\
    Offensive     &303 & 4836      \\
    Hate Inducing     & 1765       & 2399 \\
    \bottomrule
     Total     & 3189       & 14509 \\
  \end{tabular}
\end{table}

\begin{table}[H]
  \caption{Examples in the dataset}
  \label{labelleddataset}
  \centering
  \begin{tabular}{lll}
    \toprule
    \multicolumn{2}{c}{Hinglish and English Data}                   \\
    \cmidrule(r){1-3}
    Label     & HOT     & English \\
    \midrule
    Non-Offensive & \multicolumn{1}{m{3cm}}{Hum sab ghumne jaa rahe hain? http://t.} & \multicolumn{1}{m{3cm}}{We all are going outside? http://t...}     \\
    Offensive     &\multicolumn{1}{m{3cm}}{@username1 <redacted content>! Mujhe mat sikha:/} & \multicolumn{1}{m{3cm}}{@username1 <redacted content>! Do not teach me:/}     \\
    Hate Inducing     & \multicolumn{1}{m{3cm}}{<redacted content> terrorist Akbaar kill SaveWorld} & \multicolumn{1}{m{3cm}}{<redacted content> Kill terrorist Akbaar SaveWorld}     \\
    \bottomrule
    
  \end{tabular}
\end{table}

\subsection{Challenges}
The obtained data set had many challenges and thus a data preparation task was employed to clean the data and make it ready for the deep learning pipeline. The challenges and processes that were applied are stated below:
\begin{enumerate}
    \item  \textbf{Messy text messages}: The tweets had urls, punctuations, username mentions, hastags, emoticons, numbers and lots of special characters. These were all cleaned up in a preprocessing cycle 
    to clean the data. 
    \item  \textbf{Stop words}: Stop words corpus obtained from NLTK was used to eliminate most unproductive words which provide little information about individual tweets. 
    \item \textbf{Transliteration}: Followed by above two processes, we translated Hinglish tweets into English words using a two phase process
    \item \textbf{Transliteration}: In phase I, we used translation API's provided by Google translation services and exposed via a SDK, to transliteration the Hinglish messages to English messages.
    \item \textbf{Translation}: After transliteration, words that were specific to Hinglish were translated to English using an Hinglish-English dictionary. By doing this we converted the Hinglish message to and assortment of isolated words being presented in the message in a sequence that can also be represented using word to vector representation. 
    \item \textbf{Data augmentation}: Given the data set was very small with a high degree of imbalance in the labelled messages for three different classes, we employed a data augmentation technique to boost the learning of the deep network. Following techniques from the paper by Jason et al. was utilized in this setting that really helped during the training phase.Thsi techniques wasnt used in previous studies. The techniques were: 
    \begin{itemize}
        \item \textbf{Synonym Replacement (SR)}:Randomly choose \textit{n} words from the sentence that are not stop words. Replace each of these words with one of its synonyms chosen at random.
        \item \textbf{Random Insertion (RI)}:Find a random synonym of a random word in the sentence that is not a stop word. Insert that synonym into a random position in the sentence. Do this \textit{n} times.
        \item \textbf{Random Swap (RS)}:Randomly choose two words in the sentence and swap their positions. Do this \textit{n} times.
         \item \textbf{Random Deletion (RD)}:For each word in the sentence, randomly remove it with probability \textit{p}.
    \end{itemize}
    \item \textbf{Word Representation}: We used word embedding representations by Glove for creating word embedding layers and to obtain the word sequence vector representations of the processed tweets. The pre-trained embedding dimension were one of the hyperparamaters for model. Further more, we introduced another bit flag hyperparameter that determined if to freeze these learnt embedding. 
    \item \textbf{Train-test split}: The labelled dataset that was available for this task was very limited in number of examples and thus as noted above few data augmentation techniques were applied to boost the learning of the network. Before applying augmentation, a train-test split of 78\%-22\% was done from the original, cleansed data set. Thus, 700 tweets/messages were held out for testing. All model evaluation were done in on the test set that got generated by this process. The results presented in this report are based on the performance of the model on the test set. The training set of 2489 messages were however  sent to an offline pipeline for augmenting the data. The resulting training dataset was thus 7934 messages. the final distribution of messages for training and test was thus below:
    
    \begin{table}[H]
  \caption{Train-test split}
  \label{labelleddataset}
  \centering
  \begin{tabular}{lll}
    \toprule
    \multicolumn{2}{c}{Hinglish and English Data}                   \\
    \cmidrule(r){1-3}
    Label     & Train     & Test \\
    \midrule
    Non-Offensive & 3572  & 228     \\
    Offensive     &1638 & 69      \\
    Hate Inducing     & 2724       & 403 \\
    \bottomrule
     Total     & 7934       & 700 \\
  \end{tabular}
\end{table}
\end{enumerate}

\section{Model Architecture}

We tested the performance of various model architectures by running our experiment over 100 times on a CPU based compute which later as migrated to GPU based compute to overcome the slow learning progress. Our universal metric for minimizing was the validation loss and we employed various operational techniques for optimizing on the learning process. These processes and its implementation details will be discussed later but they were learning rate decay, early stopping, model checkpointing and reducing learning rate on plateau. 

\subsection{Loss function}
For the loss function we chose categorical cross entropy loss in finding the most optimal weights/parameters of the model.
Formally this loss function for the model is defined as below:

\begin{equation}
  -\frac{1}{N}{\sum}_{i=1}^{N}{\sum}_{c=1}^{C}{\textbf{1}}_{y_i \in C_c} P_{model}[y_i \in C_c]
\end{equation}

The double sum is over the number of observations and the categories respectively. While the model probability is the probability that the observation \textit{i} belongs to category \textit{c}.

\subsection{Models}
Among the model architectures we experimented with and without data augmentation were: 
\begin{itemize}
    \item \textbf{Fully Connected dense networks}: Model hyperparameters were inspired from the previous work done by Vo et al and Mathur et al.  This was also used as a baseline model but we did not get appreciable performance on such architecture due to FC networks not being able to capture local and long term dependencies. 
    \item \textbf{Convolution based architectures}: Architecture and hyperparameter choices were chosen from the past study Deon on the subject. We were able to boost the performance as compared to only FC based network but we noticed better performance from architectures that are suitable to sequences such as text messages or any timeseries data. 
    \item \textbf{Sequence models}: We used SimpleRNN, LSTM, GRU, Bidirectional LSTM model architecture to capture long term dependencies of the messages in determining the class the message or the tweet belonged to.
\end{itemize}

\newpage
Based on all the experiments we conducted below model had best performance related to metrics - Recall rate, F1 score and Overall accuracy. 
\begin{figure}[htp]
    \centering
    \includegraphics[width=15cm]{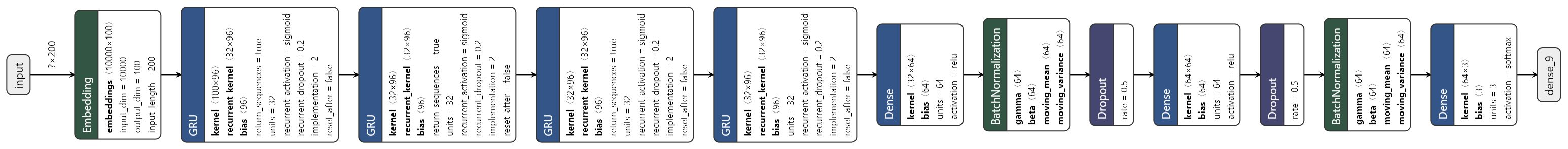}
    \caption{Deep learning network used for the modeling}
    \label{fig:Network}
\end{figure}
\subsection{Hyper parameters}
Choice of model parameters were in the above models were inspired from previous work done but then were tuned to the best performance of the Test dataset. Following parameters were considered for tuning. 
\begin{enumerate}
    \item Learning rate: Based on grid search the best performance was achieved when learning rate was set to 0.01. This value was arrived by a grid search on lr parameter. 
    \item Number of Bidirectional LSTM units: A set of 32, 64, 128 hidden activation units were considered for tuning the model. 128 was a choice made by Vo et al in modeling for Vietnamese language but with our experiments and with a small dataset to avoid overfitting to train dataset, a smaller unit sizes were considered. 
    \item Embedding dimension: 50, 100 and 200 dimension word representation from Glove word embedding were considered and the best results were obtained with 100d representation, consistent with choices made in the previous work. 
    \item Transfer learning on Embedding; Another bit flag for training the embedding on the train data or freezing the embedding from Glove was used. It was determined that set of pre-trained weights from Glove was best when it was fine tuned with Hinglish data. It provides evidence that a separate word or sentence level embedding when learnt for Hinglish text analysis will be very useful. 
    \item Number of dense FC layers.
    \item Maximum length of the sequence to be considered: The max length of tweets/message in the dataset was 1265 while average was 116. We determined that choosing 200 resulted in the best performance.  
\end{enumerate}

\section{Results}
\begin{figure}[htp]
    \centering
    \includegraphics[width=15cm]{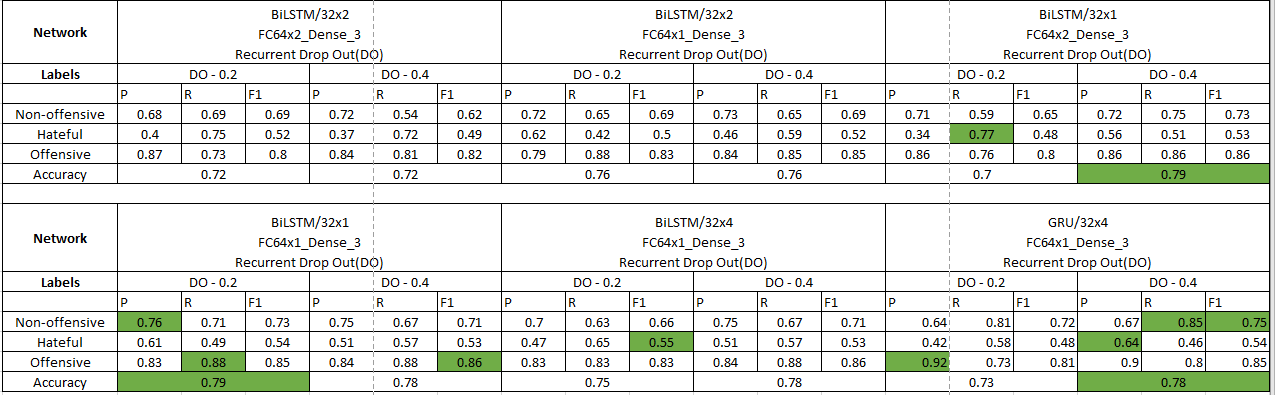}
    \caption{Results of various experiments}
    \label{fig:Network}
\end{figure}
\newpage

During our experimentation, it was evident that this is a hard problem especially detecting the hate speech, text in a code- mixed language. The best recall rate of 77 \% for hate speech was obtained by a Bidirectional LSTM with 32 units with a recurrent drop out rate of 0.2. Precision wise GRU type of RNN sequence model faired better than other kinds for hate speech detection. On the other hand for detecting offensive and non offensive tweets, fairly satisfactory results were obtained. For offensive tweets, 92 \% precision was and recall rate of 88\% was obtained with GRU versus BiLSTM based models. Comparatively, Recall of 85 \% and precision of 76 \% was obtained by again GRU and BiLSTM based models as shown and marked in the results. 

\section{Conclusion and Future work}
The results of the experiments are encouraging on detective offensive vs non offensive tweets and messages written in Hinglish in social media. The utilization of data augmentation technique in this classification task was one of the vital contributions which led us to surpass results obtained by previous state of the art Hybrid CNN-LSTM based models. However, the results of the model for predicting hateful tweets on the contrary brings forth some shortcomings of the model. The biggest shortcoming on the model based on error analysis indicates less than generalized examples presented by the dataset. We also note that the embedding learnt from the Hinglish data set may be lacking and require extensive training to have competent word representations of Hinglish text. Given this learning's, we identify that creating word embeddings on much larger Hinglish corpora may have significant results. We also hypothesize that considering alternate methods than translation and transliteration may prove beneficial.

\section*{References}
	[1] Mathur, Puneet and Sawhney, Ramit and Ayyar, Meghna and Shah, Rajiv, {\it Did you offend me? classification of offensive tweets in hinglish language},
	Proceedings of the 2nd Workshop on Abusive Language Online (ALW2)
	
	[2] Mathur, Puneet and Shah, Rajiv and Sawhney, Ramit and Mahata, Debanjan {\it Detecting offensive tweets in hindi-english code-switched language}  Proceedings of the Sixth International Workshop on Natural Language Processing for Social Media
	
	[3] Vo, Quan-Hoang and Nguyen, Huy-Tien and Le, Bac and Nguyen, Minh-Le {\it Multi-channel LSTM-CNN model for Vietnamese sentiment analysis}  2017 9th international conference on knowledge and systems engineering (KSE)
	
	[4] Hochreiter, Sepp and Schmidhuber, J{\"u}rgen {\it Long short-term memory}  Neural computation 1997
	
	[5] Sinha, R Mahesh K and Thakur, Anil {\it Multi-channel LSTM-CNN model for Vietnamese sentiment analysis}  2017 9th international conference on knowledge and systems engineering (KSE)
	
	[6] Pennington, Jeffrey and Socher, Richard and Manning, Christopher {\it Glove: Global vectors for word representation}  Proceedings of the 2014 conference on empirical methods in natural language processing (EMNLP)
	
	[7] Zhang, Lei and Wang, Shuai and Liu, Bing {\it Deep learning for sentiment analysis: A survey}  Wiley Interdisciplinary Reviews: Data Mining and Knowledge Discovery
	
	[8] Caruana, Rich and Lawrence, Steve and Giles, C Lee {\it Overfitting in neural nets: Backpropagation, conjugate gradient, and early stopping}  Advances in neural information processing systems
	
	[9] Beale, Mark Hudson and Hagan, Martin T and Demuth, Howard B {\it Neural network toolbox user’s guide}  The MathWorks Incs
	
	[10] Chollet, Fran{\c{c}}ois and others {\it Keras: The python deep learning library}  Astrophysics Source Code Library
	
	[11] Wei, Jason  and Zou, Kai {\it {EDA}: Easy Data Augmentation Techniques for Boosting Performance on Text Classification Tasks}  Proceedings of the 2019 Conference on Empirical Methods in Natural Language Processing and the 9th International Joint Conference on Natural Language Processing (EMNLP-IJCNLP)
\end{document}